\documentclass[runningheads]{llncs}
\usepackage{graphicx}
\usepackage{amsmath,amssymb} 
\usepackage{color}

\usepackage{cellspace}
\def\onedot{.\hspace{0.1cm}}

\def\ie{\emph{i.e}\onedot}

\newcommand{\PAR}[1]{\vskip0.5pt \noindent {\bf #1~}}
\newcommand{\PARbegin}[1]{\vskip1pt \noindent {\bf #1~}}

\newcommand{\setupname}[1]{\textbf{#1}}
\newcommand{\categoryname}[1]{\textit{#1}}

\begin{document}
\title{Towards Large-Scale Video Object Mining} 

\titlerunning{Towards Large-Scale Video Video Object Mining}
%
\author{
Aljo\u{s}a O\u{s}ep\textsuperscript{*}\inst{1} \and
Paul Voigtlaender\textsuperscript{*}\inst{1} \and
Jonathon Luiten\inst{1} \and 
Stefan Breuers\inst{1} \and
Bastian Leibe\inst{1}
}
\authorrunning{A. O\u{s}ep\textsuperscript{*} and P. Voigtlaender\textsuperscript{*} and J. Luiten and S. Breuers and B. Leibe}
%
\institute{
Computer Vision Group, RWTH Aachen University, Germany\\
\email{lastname@vision.rwth-aachen.de}
}
\maketitle              
\vspace{-10mm}
\begin{abstract}
We propose to leverage a generic object tracker in order to perform object mining in large-scale unlabeled videos, captured in a realistic automotive setting.
We present a dataset of more than $360'000$ automatically mined object tracks from $10+$ hours of video data ($560'000$ frames) and propose a method for automated novel category discovery and detector learning. In addition, we show preliminary results on using the mined tracks for object detector adaptation.
%
%
\end{abstract}

\vspace{-10mm}
\section{Introduction}
\vspace{-3mm}
\footnotetext{\textsuperscript{*} Equal contribution.}
Deep learning has revolutionized the way research is being performed in computer vision and in particular in autonomous driving.
However, deep learning requires huge quantities of annotated training data, which are very costly to obtain.
This problem is of particular relevance in
mobile robotics, where future intelligent agents will encounter a multitude of relevant and novel object categories, most of which are not captured by today's detectors. These object categories appear in the long tail of the object distribution making it hard to acquire sufficient training examples.
In order to react to such situations, future robots need the capability to
automatically identify these objects and update existing object detectors.

In this paper, we propose a method for large-scale video-object mining.
We apply our method to two large datasets (KITTI Raw~\cite{Geiger13IJRR} and Oxford RobotCar~\cite{Maddern17IJRR}) for autonomous driving, comprising altogether roughly 10 hours of video consisting of more than $560'000$ frames. From this data, we extract  more than 360'000 generic object tracks without any human supervision and covering both known and unknown object categories. These can serve as a basis for research in object detector domain adaptation and new detector learning based on the discovered object categories.
We manually annotated two subsets of $6'000$ and $12'000$ of these tracks into 36 categories to serve as a benchmark set for object discovery in-the-wild. Importantly, these annotations include many categories not contained in the COCO dataset~\cite{Lin14ECCV} commonly used for training object detectors.

%
%
\vspace{-4mm}
\section{Method}
\vspace{-3mm}
Towards the goal of detector domain adaptation and novel detector learning, we need to be able to mine object samples from unlabeled video for both, \textit{known} 
and \textit{unknown} object categories.
For this task, we propose to leverage temporal information and prior knowledge about common object categories. By forming object tracks from image-level object proposals (using an adaptation of the generic stereo-based tracker from \cite{Osep18ICRA}) we reduce the object candidate space by a large margin and suppress noise and clutter in image-level object proposals.
Recognition of certain common object types (using the classifier component of a Faster R-CNN~\cite{Ren15NIPS}) further reduces the proposal space and helps resolve ambiguities.
Tracks are thus automatically labeled by the recognized category type (\ie as one of the COCO~\cite{Lin14ECCV} categories) or as \textit{unknown} object track.

\PAR{Novel Object Discovery.}
After running the tracker we obtain a reduced set of object tracks, each is
marked as either \textit{known} or \textit{unknown}.
In order to train novel object detectors, we need to find patterns in the unknown object set.
We propose to do this using clustering methods. In particular, we perform clustering in a learned embedding space using HDBSCAN~\cite{Campello15TKDD} due to its scalability and ability to deal with outliers.
One of the central questions in clustering is how to define a distance measure between data points. Tracks are defined by a collection of image crops, representing the appearance of the tracked object over time. When applying the embedding network on tracks, we extract a representative embedding vector for each track by taking the embedding vector of the crop closest to the mean of the embedding vectors of the track`s image crops.
For obtaining image-crop embeddings, a simple yet surprisingly effective method is to utilize a pre-trained network to extract features from its internal activations.
Another possibility is to make use of feature embedding learning using the triplet loss~\cite{Weinberger09JMLR}. The idea here is to use a labeled dataset to learn a feature embedding in which images of the same class have a small distance and images of different classes are far away. We experimentally compare both strategies.

\PAR{Detector Adaptation and Learning Using Noisy Track Data.}
We utilize Faster R-CNN~\cite{Ren15NIPS} for robust detector fine-tuning and novel detector learning using the mined tracks.
When working with manually annotated data, using dense anchor boxes coverage as potential training examples~\cite{Ren15NIPS} is reasonable, but when obtaining annotations automatically by tracking, this strategy would be sensitive to missed targets and tracking errors.
We propose to select only confident anchors boxes - those that have IoU overlap of at least $50\%$ with either a \textit{known} object track (fine-tuning) or a cluster-member track (new detector learning).
To obtain negatives, we exploit simple geometric knowledge.
The tracker provides a 3D ground-plane estimate, which we use to mask the ground-pixels and the area that spans more than 2.5 m above the ground. We assume no objects in this area and retain anchor boxes with sufficient overlap to this region as negatives.
%
%
By using the clustering results instead of the Faster R-CNN (track) classification, we can learn detectors for previously unseen categories.

\vspace{-4mm}
\section{Experiments}
\label{sec:exp}
\vspace{-3mm}
\PARbegin{Track Collections.}
For our experiments, we used a subset of 1.18 h of KITTI Raw~\cite{Geiger13IJRR} and a subset of 9 h of video of the Oxford Robotcar dataset~\cite{Maddern17IJRR}.
Tab.~\ref{tab:detector-finetuning} \textit{(left)} displays a short summary of the tracks mined for both datasets.
Using generic object tracking, we are able to reduce the number of object candidates to a manageable level and achieve a significant compression factor per image and an even greater compression factor on the sequence level.
%
For the track analysis and clustering evaluation, we manually annotated a subset of mined tracks. 
%
%
We label each track as one of 36 categories or mark it as a valid \textit{unknown} object. Tracks that diverge from the tracked object are marked as a tracking error. In particular, the results show that only between $6.4\%$ (Oxford Track Collection, OTC) and $9.3\%$ (KITTI Track Collection, KTC) of generic object tracks are affected by tracking errors.

\begin{figure}[t]
    \begin{center}
       \includegraphics[width=0.3\linewidth]{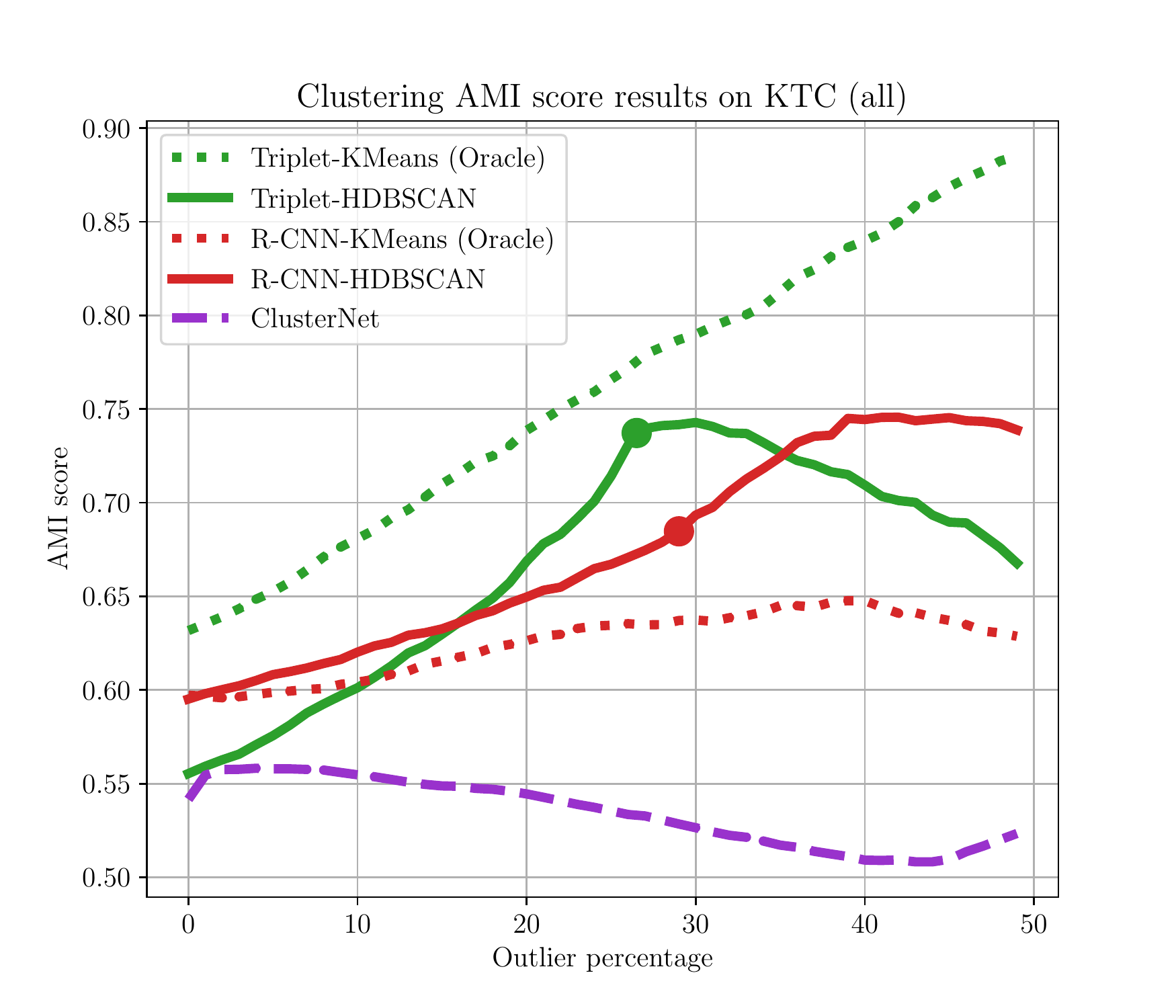}
       \includegraphics[width=0.3\linewidth]{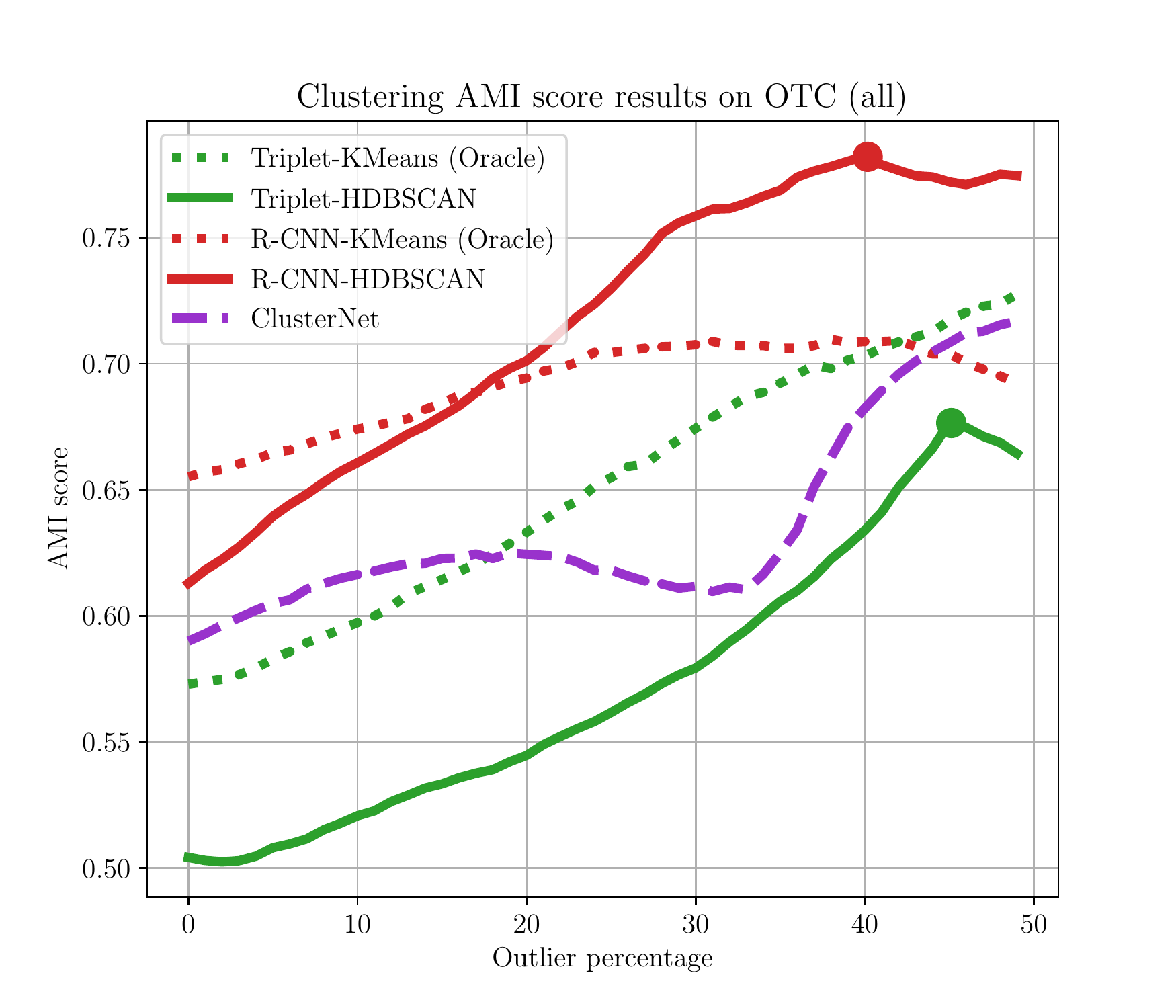}
       \includegraphics[width=0.3\linewidth]{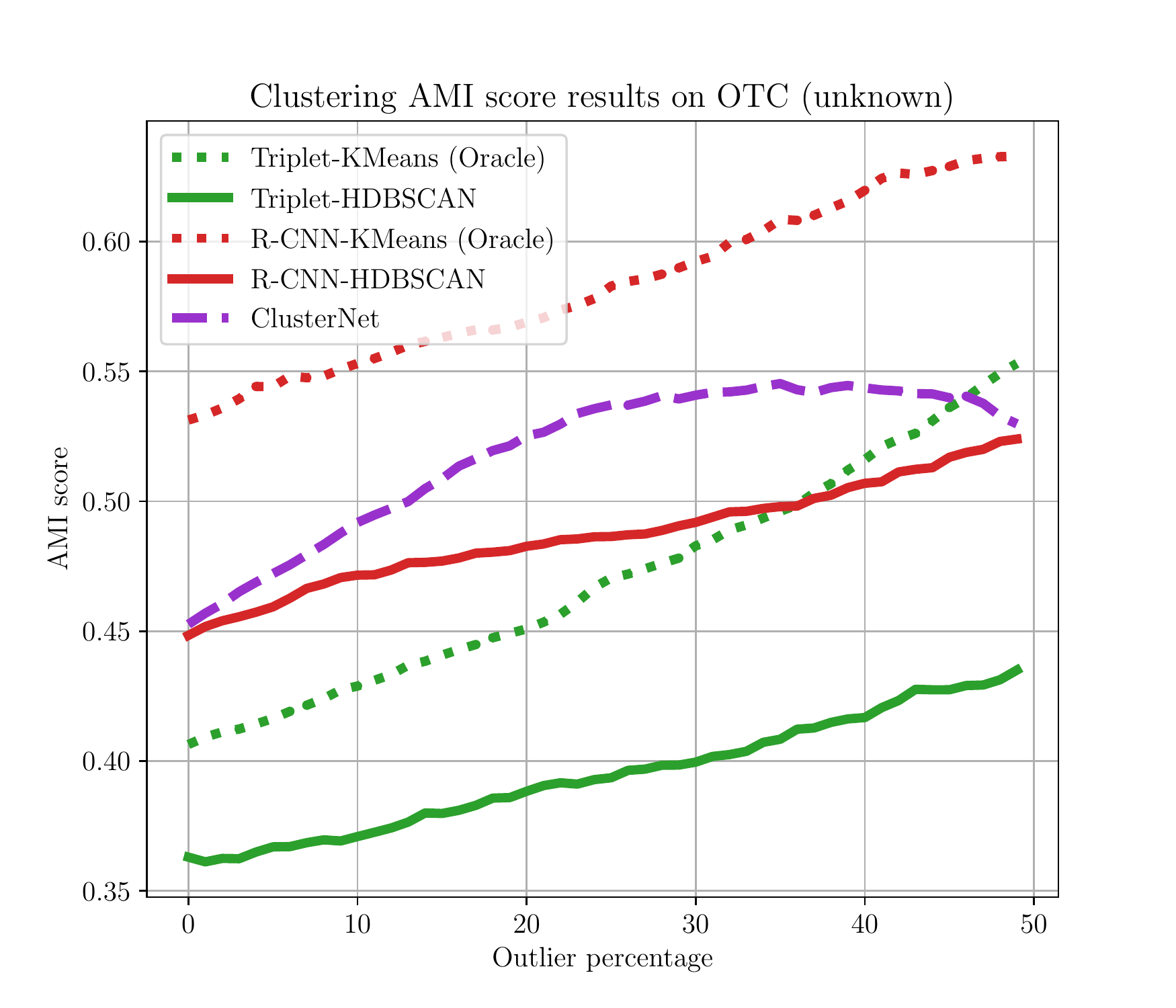}
    \end{center}
    \vspace{-25pt}
       \caption{Clustering results measured by AMI. Circle markers represent the automatically selected fraction of outliers by HDBSCAN.}
   \label{fig:clustering-eval}
\end{figure}

\begin{figure}[t]
\begin{center}
   \includegraphics[width=1.0\linewidth]{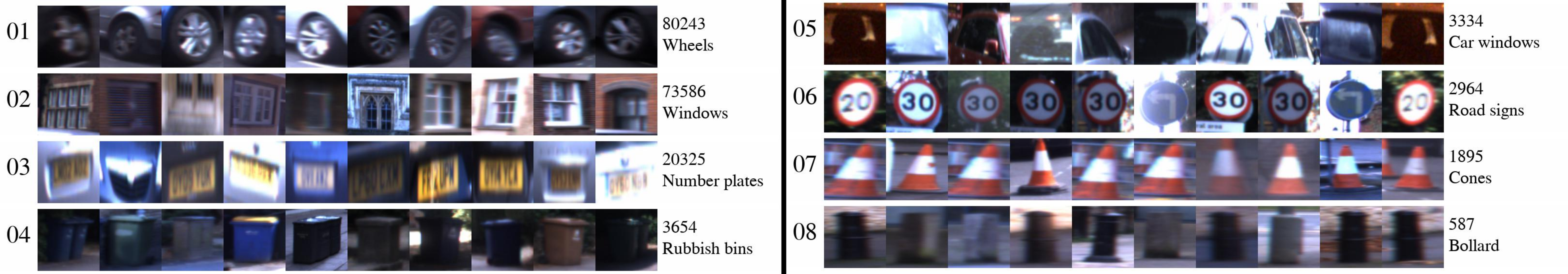}
   \vspace{-22pt}
   \caption{\label{fig:oxford-clusters} Subset of newly discovered categories (clustering results) on Oxford. The numbers on the right hand side indicate the number of tracks in each cluster.}
\end{center}
\end{figure}

\PAR{Object Discovery.}
We compare a number of two-stage (embedding-clustering) methods and ClusterNet~\cite{Hsu16ARXIV}. We extract two different embeddings, last layer Faster R-CNN features and a $128$D triplet loss~\cite{Weinberger09JMLR} embedding network trained to separate the 80 COCO classes. We use two clustering methods on these embeddings: KMeans (with ground-truth number of clusters as an upper bound) and HDBSCAN.
ClusterNet (with 50 cluster labels) is trained using the output of a Similarity Prediction Network (SPN), which is first trained on COCO to predict whether two crops belong to the same class.
Fig.~\ref{fig:clustering-eval} shows results on both datasets comparing the Adjusted Mutual Information (AMI) to the outlier fraction (based on distance to cluster center), for all annotated categories and for non-COCO categories only.
In Fig.~\ref{fig:oxford-clusters} we show qualitatively that we were able to discover new object categories.

\newcommand\hoxf{1.2}
\newcommand\hkit{1}
\begin{figure}[t]
\begin{center}
    \label{fig:oxford-qualitative}
   \includegraphics[height=\hoxf cm]{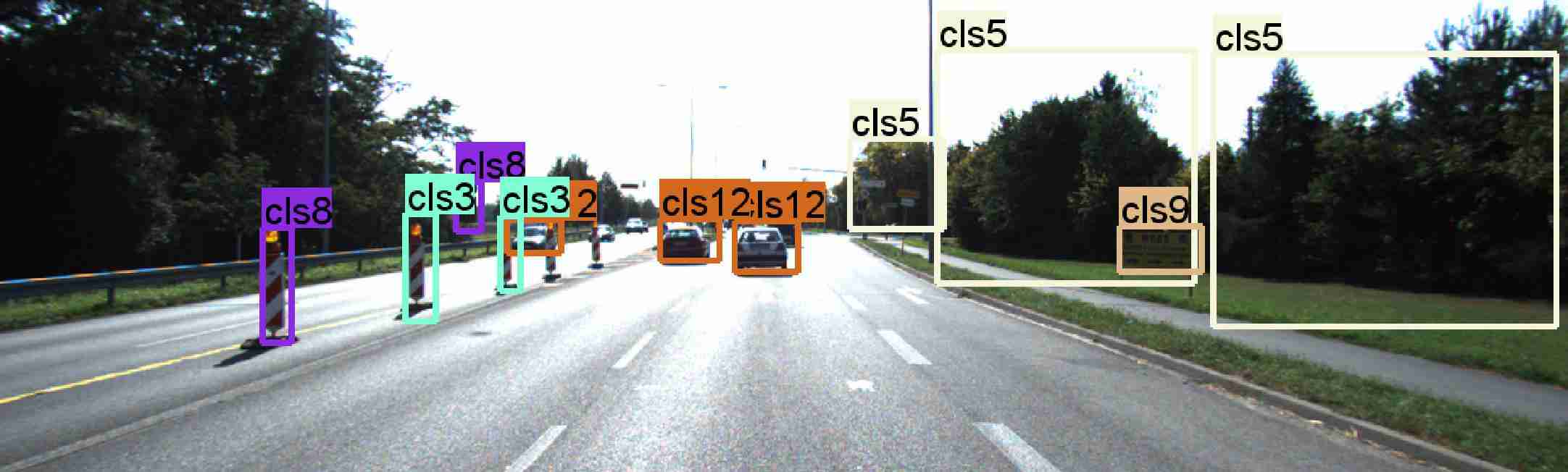}
   \includegraphics[height=\hoxf cm]{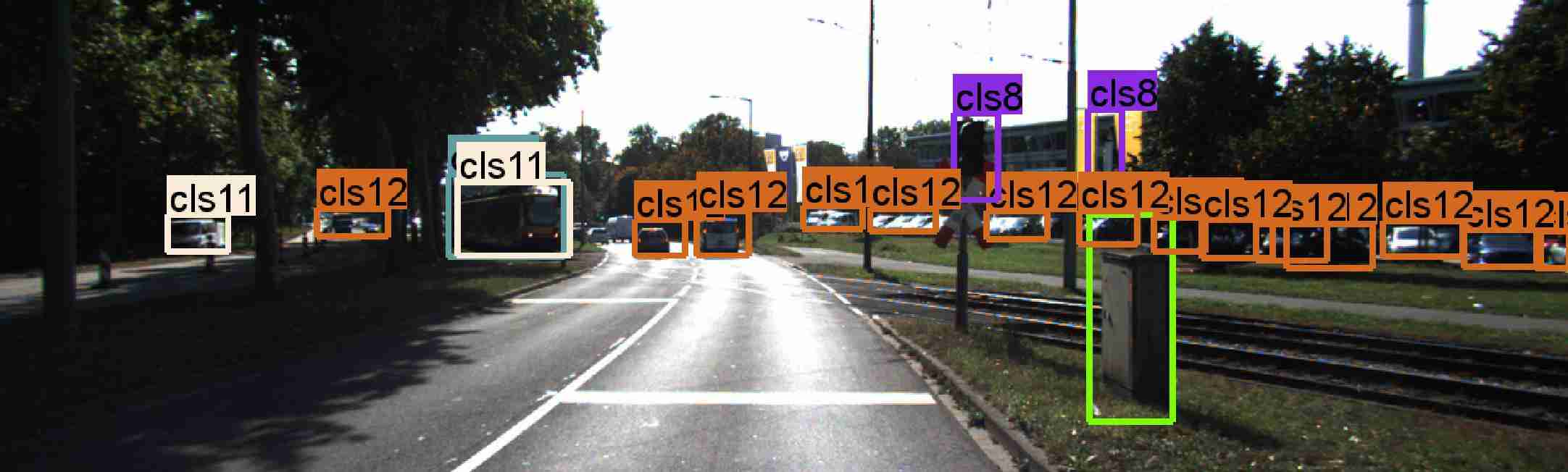}
   \includegraphics[height=\hoxf cm]{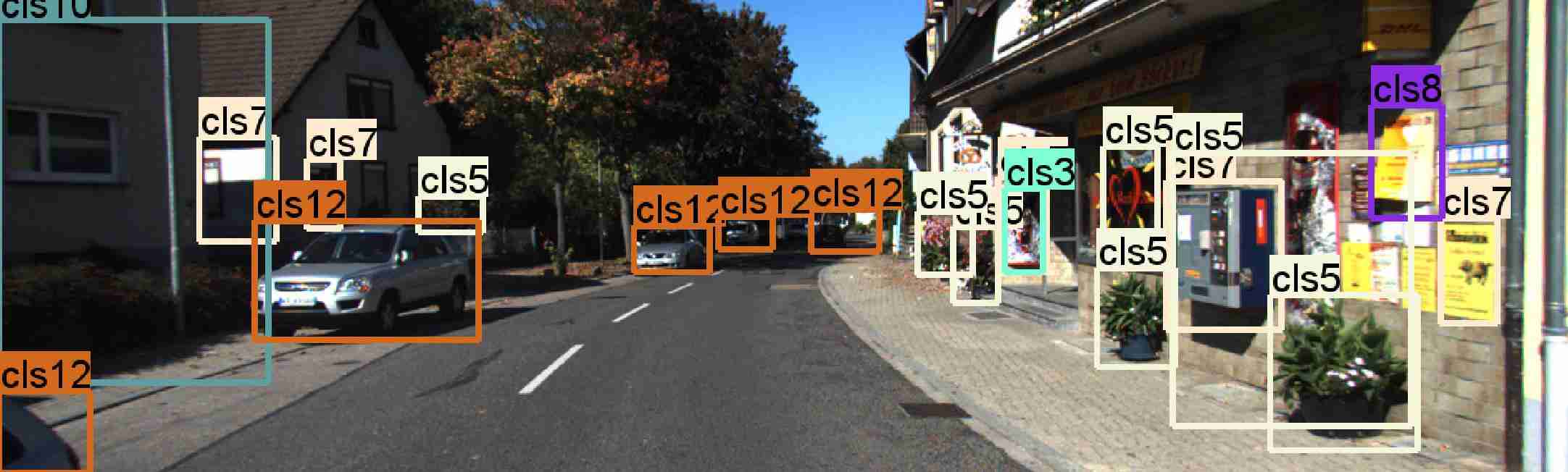}\\
   \includegraphics[height=\hoxf cm]{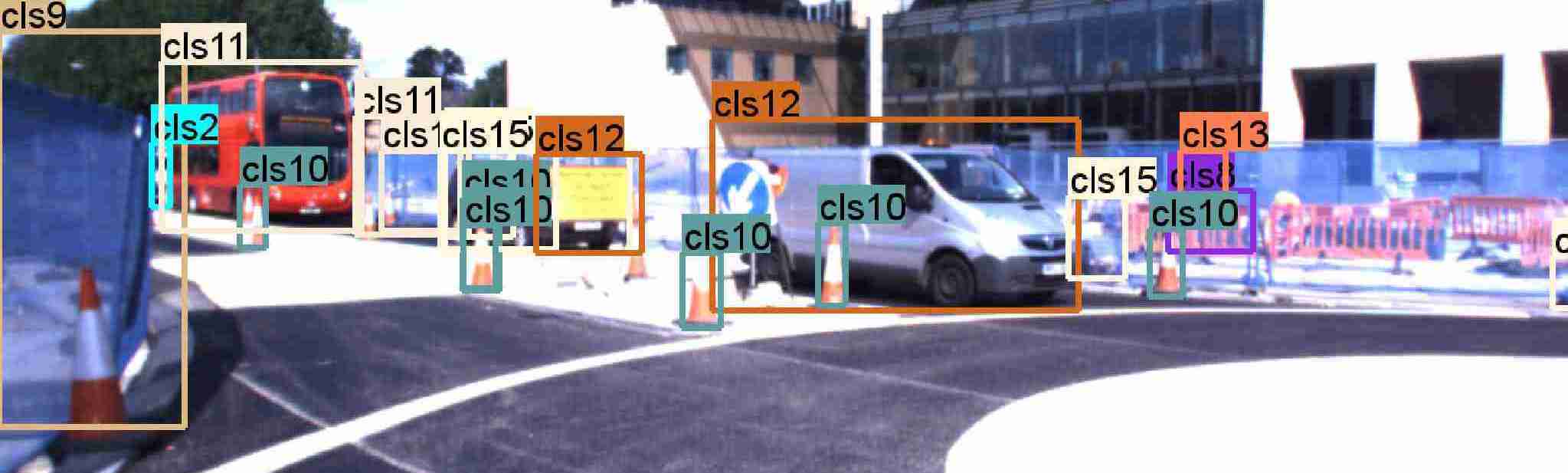}
   \includegraphics[height=\hoxf cm]{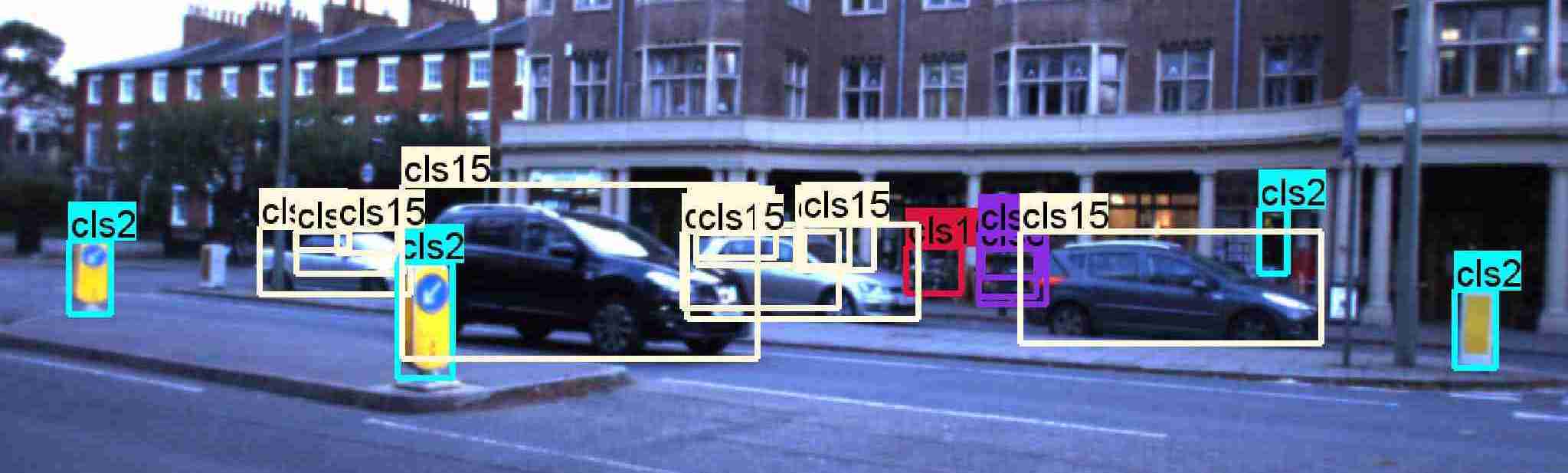}
   \includegraphics[height=\hoxf cm]{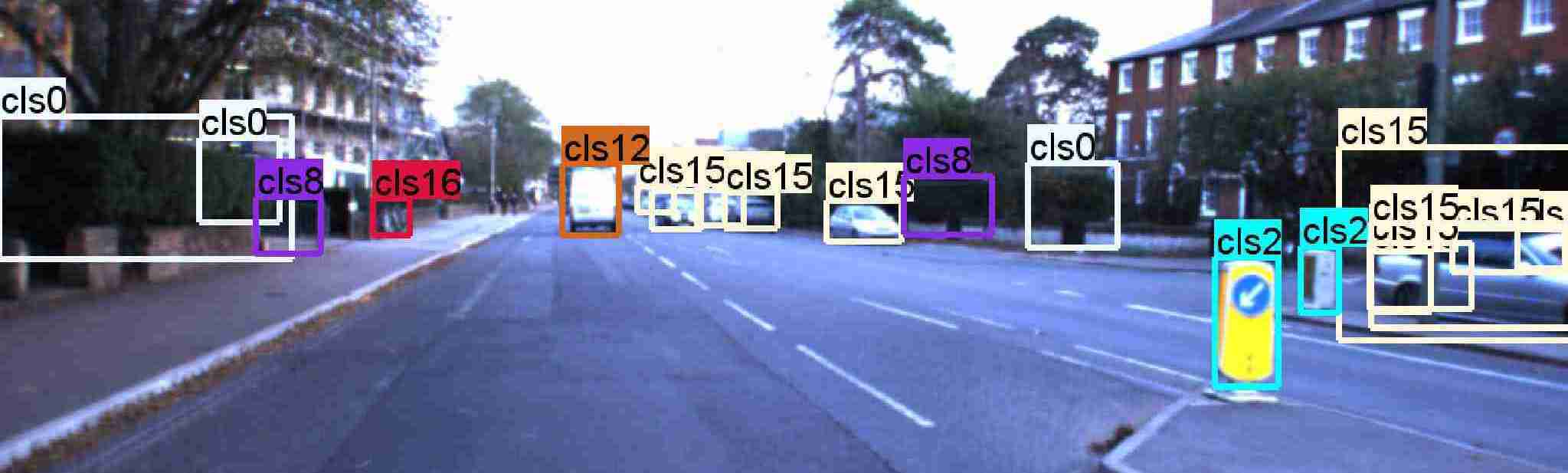}
   \vspace{-22pt}
   \caption{\label{fig:object_detect} Proof-of-concept object detection results, trained using clustered tracks.}
\end{center}
\end{figure}

\PAR{Detector Adaptation and Learning.}
\begin{table}[t]
\tiny
  \begin{center}
          \begin{tabular}{l|r|r|r}
       & \textbf{KTC} & \textbf{OTC}\\ 
        \hline
        Frames & 42,407 & 521,500 \\
        Duration (h) & 1.18 & 9.06\\
        Proposals (all) & 4,240,700 & 52,150,000\\
        Tracks (total) & 8,005 & 359,503\\
        Tracks (labeled) & 8,005 & 12,308\\
        Tracks (unk.) & 1,190 & 4,198\\
        Tracking Errors & 745 & 787\\
        \end{tabular}
        \quad
    \begin{tabular}{l|r|r|r}
      Fine-tuned on & AP car & AP pedestrian & AP cyclist\\ 
        \hline
        - & 69.9 & 58.5 & 0\\
        KITTI GT & 79.9 & 74.9 & 60.7 \\
        KITTI det. & 69.2 & 59.4 & 0\\
        KITTI tracks & 61.2 & 65.9 & 19.3 \\
        KITTI tracks + COCO & 67.2 & 68.9 & 22.6 \\
         KITTI tracks, no subs. & 67.9 & 55.4 & 24.3 \\
        \end{tabular}
    \vspace{2pt}
    \caption{\textit{(Left)} Track mining statstics. \textit{(Right)} Results of detector fine-tuning on KITTI Raw (avg. precision (AP, \%)). }
    \vspace{-5mm}
    \label{tab:detector-finetuning}
  \end{center}
\end{table}
The first line in Tab.~\ref{tab:detector-finetuning} \textit{(right)} indicates the performance of the COCO pre-trained detector.
The setup \setupname{KITTI GT}
provides a strong baseline which was trained on manually annotated data.
For the baseline \setupname{KITTI det.} we use for training confident predictions (threshold of $0.3$) of COCO pre-trained detector on KITTI Raw. Fine-tuning the detector in this way does not significantly change the detector performance.
When fine-tuning only on the \setupname{KITTI tracks}, the performance for \categoryname{cars} degrades, since the COCO pre-trained detector was already very strong on cars.
For \categoryname{pedestrians}, however, we can significantly improve over the COCO baseline from $58.5\%$ to $65.9\%$. For \categoryname{cyclists} we merged during the training tracks recognized as \categoryname{bicycle} and \categoryname{person} when their spatial distance was smaller than one meter.
To prevent the degradation of performance for \categoryname{cars}, we propose to train on the mined tracks and COCO jointly.
When not using proposed anchor subselection (\setupname{KITTI tracks, no subs.}), the detection performance for pedestrians degrades significantly.
Fig.~\ref{fig:oxford-qualitative} shows proof-of-concept qualitative results for training detectors on the automatically mined clusters. 
%
%
\vspace{-4mm}
\section{Conclusion}
\vspace{-3mm}
This work is an initial study about learning from unlabeled data by automatically extracting generic object tracks. We show preliminary results on object detector fine-tuning and new detector learning based on newly-discovered, previously unseen categories.
We believe that this work is a starting point for exciting new research with a large potential for further exploiting unlabeled video data.
\footnotesize {\PAR{Acknowledgments:} This project was funded, in parts, by ERC Consolidator
Grant DeeVise (ERC-2017-COG-773161).}
%
\vspace{-5mm}
%
%
%
%
\bibliographystyle{splncs04}
\bibliography{abbrev_short,best_paper}
\end{document}